\documentclass{article} 
\usepackage{nips14submit_e,times}

\usepackage{hyperref}
\usepackage{url}
\usepackage[ruled]{algorithm}
\usepackage{algpseudocode}
\usepackage{amsmath}
\usepackage{amssymb}
\usepackage{graphicx}
\usepackage{wrapfig}
\usepackage{caption}
\usepackage{subcaption}
\usepackage{xspace}
\usepackage{titlesec}

\titlespacing*{\section}{0pt}{2.0ex plus 1ex minus .2ex}{1.0ex plus .2ex}

\newcommand{\R}{\mathbb{R}}

\newcommand{\mat}[1]{{\mathbf{#1}}}
\newcommand{\randn}{\operatorname{randn}}
\newcommand{\orth}{\operatorname{orth}}
\newcommand{\chol}{\operatorname{chol}}
\newcommand{\svd}{\operatorname{svd}}
\newcommand{\RandomizedCCA}{RandomizedCCA\xspace}

\algdef{SE}[DATAPASS]{Datapass}{EndDatapass}{\textbf{data pass}}{\textbf{end data pass}}

\DeclareMathOperator*{\maximize}{maximize\;}
\DeclareMathOperator*{\subjectto}{subject\ to\;}
\newcommand{\Tr}{\mathrm{Tr}}

\title{A Randomized Algorithm for CCA}
\author{
Paul Mineiro \\
Microsoft CISL \\
\texttt{pmineiro@microsoft.com}
\And
Nikos Karampatziakis \\
Microsoft CISL \\
\texttt{nikosk@microsoft.com}
}

\nipsfinalcopy 

\begin{document}

\maketitle

\begin{abstract}
We present \RandomizedCCA, a randomized algorithm for computing canonical
analysis, suitable for large datasets stored either out of core or on a
distributed file system.  Accurate results can be obtained in as few as two
data passes, which is relevant for distributed processing frameworks in
which iteration is expensive (e.g., Hadoop).  The strategy also provides
an excellent initializer for standard iterative solutions.
\end{abstract}

\section{Introduction}

Canonical Correlation Analysis (CCA) is a fundamental statistical
technique for characterizing the linear relationships
between two\footnote{CCA can be extended to more than two
views, but we don't pursue this here.} multidimensional
variables.\footnote{Furthermore, nonlinear relationships between
variables can be uncovered using kernel CCA, or, for large-scale
data sets, a primal approximation e.g., with randomized feature
maps\cite{lopez2014randomized} or the Nystr\"om method.} First
introduced in 1936 by Hotelling\cite{hotelling1936relations},
it has found numerous applications.  For the machine learning
community, more familiar applications include learning with
privileged information\cite{vapnik2009new}, semi-supervised
learning\cite{chaudhuri2009multi,mcwilliams2013correlated},
monolingual\cite{dhillon2011multi} and
multilingual\cite{faruqui2014improving} word representation
learning, locality sensitive hashing\cite{gong2013iterative} and
clustering\cite{blaschko2008correlational}.  Because these applications
involve unlabeled or partially labeled data, the amount of data available
for analysis can be vast, motivating the need for scalable approaches.

\section{Background}

Given two view data, CCA finds a projection of each view into a common
latent space which maximizes the cross-correlation, subject to each view
projection having unit variance, and subject to each projection dimension
being uncorrelated with other projection dimensions.  In matrix form,
given two views $\mat{A} \in \R^{n \times d_a}$ and $\mat{B} \in \R^{n
\times d_b}$, the CCA projections $\mat{X}_a \in \R^{d_a \times k}$
and $\mat{X}_b \in \R^{d_b \times k}$ are the solution to
\begin{align}
\maximize_{ \mat{X}_a , \mat{X}_b }& \Tr \left( \mat{X}_a^\top \mat{A}^\top \mat{B} \mat{X}_b  \right), \nonumber \\
\subjectto& \mat{X}_a^\top \mat{A}^\top \mat{A} \mat{X}_a = n \mat{I}, \label{eqn:awhite} \\
\;& \mat{X}_b^\top \mat{B}^\top \mat{B} \mat{X}_b = n \mat{I}. \label{eqn:bwhite} 
\end{align}
The KKT conditions, expressed in terms of the QR-decompositions $\mat{Q}_a \mat{R}_a = \mat{A}$ and $\mat{Q}_b \mat{R}_b = \mat{B}$,  lead to the following multivariate eigenvalue problem\cite{chu1993multivariate}
\begin{equation} \label{eqn:cca}
\left( \begin{array}{cc} 0 & \mat{Q}_a^\top \mat{Q}_b \\ \mat{Q}_b^\top \mat{Q}_a & 0 \end{array} \right) \left( \begin{array}{c} \mat{V}_a \\ \mat{V}_b \end{array} \right) = \left( \begin{array}{c} \mat{V}_a \\ \mat{V}_b \end{array} \right)  \mat{\Lambda},
\end{equation}
subject to $\mat{V}_a^\top \mat{V}_a = \mat{I}$, $\mat{V}_b^\top \mat{V}_b = \mat{I}$, $\mat{V}_a \mat{R}_a = \mat{X}_a$, and $\mat{V}_b \mat{R}_b = \mat{X}_b$.

Equation \eqref{eqn:cca} leads to several solution strategies.
For moderate sized design matrices, an SVD of $\mat{Q}_a^\top \mat{Q}_b$
directly reveals the solution in the $\mat{V}_{a,b}$ coordinate system
\cite{bjorck1973numerical}.  The transformation from $\mat{V}_{a,b}$
to $\mat{X}_{a,b}$ can be obtained from either the SVD
or QR-decompositions of $\mat{A}$ and $\mat{B}$.

For larger design matrices lacking special structure, SVD and
QR-decompositions are prohibitively expensive, necessitating other
techniques.  Large scale solutions are possible via Horst iteration
\cite{chu1993multivariate}, the analog of orthogonal power iteration
for the multivariate eigenvalue problem, in which each block of
variables is individually normalized following matrix multiplication
\cite{zhang2011computing}.  For CCA, the matrix multiplication step of
Horst iteration can be done directly in the $\mat{X}_{a,b}$ coordinate
system via solving a least-squares problem.  Furthermore, the least
squares solutions need only be done approximately to ensure convergence
\cite{lu2014large}.  Unfortunately, Horst iteration still requires many
passes over the data for good results.

\section{Algorithm}

\begin{algorithm}[t]
  \caption{Randomized CCA}
  \begin{algorithmic}[1]
    \Statex

    \Function{RandomizedCCA}{$k, p, q, \lambda_a, \lambda_b, \mat{A} \in \R^{n \times d_a}, \mat{B} \in \R^{n \times d_b}$}
      \State \texttt{//  Randomized range finder for $\mat{A}^\top \mat{B}$} \label{lin:rangefindstart}
      \State $\mat{Q}_a \leftarrow \randn(d_a, k+p)$ \Comment{Gaussian suitable for sparse $\mat{A}$, $\mat{B}$}
      \State $\mat{Q}_b \leftarrow \randn(d_b, k+p)$ \Comment{Structured randomness suitable for dense $\mat{A}$, $\mat{B}$}
      \For{$i \in \{ 1, \ldots, q \}$} 
        \Datapass
          \State $\mat{Y}_a \leftarrow \mat{A}^\top \mat{B} \mat{Q}_b$
          \State $\mat{Y}_b \leftarrow \mat{B}^\top \mat{A} \mat{Q}_a$
        \EndDatapass
        \State $\mat{Q}_a \leftarrow \orth(\mat{Y}_a)$
        \State $\mat{Q}_b \leftarrow \orth(\mat{Y}_b)$
      \EndFor \label{lin:rangefindend}
      \State \texttt{// Final optimization over bases $\mat{Q}_a, \mat{Q}_b$}
      \Datapass 
        \State $\mat{C}_a \leftarrow \mat{Q}_a^\top \mat{A}^\top \mat{A} \mat{Q}_a$ \Comment{$\mat{C}_a \in \R^{(k+p)\times(k+p)}$ is ``small''}
        \State $\mat{C}_b \leftarrow \mat{Q}_b^\top \mat{B}^\top \mat{B} \mat{Q}_b$ \Comment{Similarly for $\mat{C}_b$, $\mat{F}$}
        \State $\mat{F} \leftarrow \mat{Q}_a^\top \mat{A}^\top \mat{B} \mat{Q}_b$ 
      \EndDatapass   
      \State $\mat{L}_a \leftarrow \chol(\mat{C}_a + \lambda_a \mat{Q}_a^\top \mat{Q}_a)$ \Comment{$\mat{Q}_a \mat{L}_a^{-1} = (\mat{A}^\top \mat{A} + \lambda_a \mat{I})^{-1/2} \mat{Q}_a$} \label{lin:reducestart}
      \State $\mat{L}_b \leftarrow \chol(\mat{C}_b + \lambda_b \mat{Q}_b^\top \mat{Q}_b)$ \Comment{$\mat{Q}_b \mat{L}_b^{-1} = (\mat{B}^\top \mat{B} + \lambda_b \mat{I})^{-1/2} \mat{Q}_b$}
      \State $\mat{F} \leftarrow \mat{L}_a^{-\top} \mat{F} \mat{L}_b^{-1}$ \label{lin:fmult}
      \State $(\mat{U}, \Sigma, \mat{V}) \leftarrow \svd(\mat{F}, k)$ \label{lin:reduceend}
      \State $\mat{X}_a \leftarrow \sqrt{n} \mat{Q}_a \mat{L}_a^{-1} \mat{U}$
      \State $\mat{X}_b \leftarrow \sqrt{n} \mat{Q}_b \mat{L}_b^{-1} \mat{V}$
      \State \Return $(\mat{X}_a, \mat{X}_b, \Sigma)$
    \EndFunction
  \end{algorithmic}
  \label{alg:randcca}
\end{algorithm}

Our proposal is \RandomizedCCA outlined in Algorithm~\ref{alg:randcca}.
For ease of exposition, we elide mean shifting of each design matrix,
which is a rank one update, and can be done in $O(d_a+d_b)$ extra space
without introducing additional data passes and preserving sparsity.
Line numbers \ref{lin:rangefindstart} through~\ref{lin:rangefindend}
constitute a standard randomized range finder~\cite{halko2011finding} with
power iteration for the left and right singular spaces of $\mat{A}^\top
\mat{B}$.  If we consider $\mat{Q}_a$ and $\mat{Q}_b$ as providing a
$\tilde k$ rank approximation to the top range of $\mat{A}^\top \mat{B}$,
then analysis of randomized range finding indicates $
\mathbb{E}\left\| \mat{A}^\top \mat{B} - \mat{Q}_a \mat{Q}_a^\top \mat{A}^\top \mat{B} \right\| \leq \left[ 1 + 4 \frac{\sqrt{k + p}}{p - \tilde k - 1} \sqrt{n} \right]^{1/q} \sigma_{\tilde k}, 
$ and analogously for $\mat{Q}_b$\cite{halko2011finding}.  Intuition
about the relevant value of $\tilde k$ can be determined by considering
the effect of regularization.  To prevent overfitting, equations
\eqref{eqn:awhite} and \eqref{eqn:bwhite} are regularized with $\lambda_a
\mat{I}$ and $\lambda_b \mat{I}$ respectively, hence the canonical
correlations possible in the space orthogonal to the top $\tilde k$ range
of $\mat{A}^\top \mat{B}$ are at most $\sigma_{\tilde k} / \sqrt{\lambda_a
\lambda_b}$.  When this quantity is below the $k^\mathrm{th}$ canonical
correlation, the top $\tilde k$ range of $\mat{A}^\top \mat{B}$ is the
only relevant subspace and the question then becomes the extent to which
the randomized range finder is approximating this space well.

In practice $\tilde k$ is unknown and thus relative to $k$, \RandomizedCCA
effectively requires large amounts of oversampling (e.g., $p = 1000$)
to achieve good results.  Nonetheless, when iterations over the data
are expensive, this level of oversampling can be more computationally
attractive than alternative approaches.  This is because typically CCA
is used to find a low dimensional embedding (e.g., $k = 50$), whereas the
final exact SVD and Cholesky factorizations in lines \ref{lin:reducestart}
through \ref{lin:reduceend} can be done using a single commodity machine
as long as $k+p \lesssim 10000$.  Therefore there is computational
headroom available for large oversampling.  Ultimately the binding
constraint is the utility of storing $\mat{Q}_{a,b}$ and $\mat{Y}_{a,b}$
in main memory.

\section{Experimental Results}

Europarl is a collection of simultaneous translated documents extracted
from the proceedings of the European parliament \cite{koehn2005europarl}.
Multilingual alignment is available at the individual sentence level. We
used a single random 9:1 split of sentences into train and test sets for
all experiments.  We processed each sentence into a fixed dimensional
representation using a bag of words representation composed with
inner-product preserving hashing \cite{weinberger2009feature}.  For these
experiments we used $2^{19}$ hash slots.\footnote{The hashing strategy
generates a feature space in which many features never occur. To reduce
memory requirements, we lazily materialize the rows of $\mat{Y}_{a,b}$
and $\mat{Q}_{a,b}$.}  We used English for the $\mat{A}$ design matrix
and Greek for the $\mat{B}$ design matrix, resulting in $n = 1,235,976$
and $d_a = d_b = 2^{19}$.  Note the ultimate embedding produced by this
procedure is the composition of the hashing strategy with the projections
found by \RandomizedCCA.

\captionsetup[figure]{font=scriptsize}
\begin{wrapfigure}{r}{0.45\textwidth} 
\vspace{-116pt}
  \begin{center}
    \noindent\makebox[0.45 \textwidth]{%
    \includegraphics[width=0.62\textwidth]{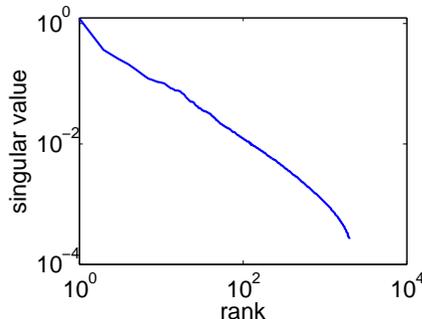}
    }
    \vspace{-115pt}
    \caption{Spectrum of $(1/n) \mat{A}^\top \mat{B}$.}
    \label{fig:atopbspectrum}
  \end{center}
  \vspace{-30pt}
\end{wrapfigure}
\clearcaptionsetup{figure}
The top-2000 spectrum of $(1/n)\mat{A}^\top \mat{B}$, as estimated by
two-pass randomized SVD, is shown in figure \ref{fig:atopbspectrum}.
This provides some intuition as to why the top range of $\mat{A}^\top
\mat{B}$ should generate an excellent approximation to the optimal
CCA solution, as the spectrum exhibits power-law decay and ultimately
decreases to a point which is comparable to a plausible regularization
parameter setting.

\begin{figure}[t]
\centering
\begin{subfigure}[b]{0.4 \textwidth}
\vspace{-120pt}
\noindent\makebox[\textwidth]{%
\includegraphics[width=1.8 \textwidth]{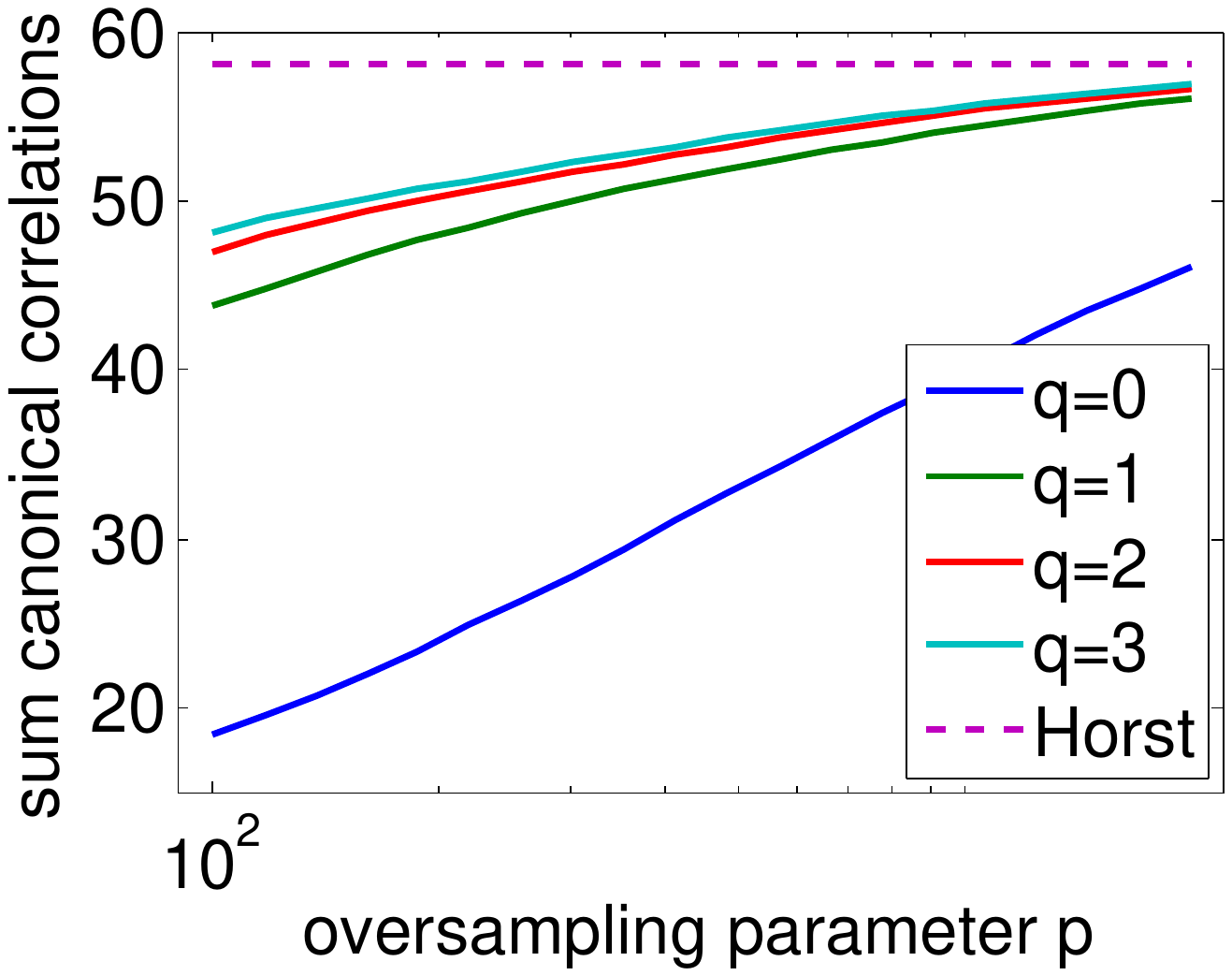}
}
\vspace{-120pt}
\caption{$\frac{1}{n} \Tr \left( \mat{X}_a \mat{A}^\top \mat{B} \mat{X}_b \right)$
for \RandomizedCCA as $q$ and $p$ are varied, with $k = 60$.  The dashed
line is the result of running Horst iteration for 120 passes
over the data.}
\label{fig:ccahyperresults}
\end{subfigure}
\hspace{0.05 \textwidth}
\begin{subtable}[b]{0.45 \textwidth}
\begin{minipage}{\textwidth}
\vspace{-200pt}
\begin{tabular}{c|c|c|c|c}
q & p & Train & Test & time (s) \\ \hline
0 & 910 & 38.942 & 38.797 & \rule{0pt}{2ex} 190 \rule{0pt}{2ex} \\
0 & 2000 & 46.095 & 45.891 & 463 \\
1 & 910 & 53.934 & 53.835 & 334 \\
1 & 2000 & 56.054 & 55.656 & 770 \\
2 & 910 & 55.017 & 54.782 & 484 \\
2 & 2000 & 56.666 & 56.528 & 1186 \\ 
3 & 910 & 55.386 & 54.991 & 637 \\ 
3 & 2000 & 56.833 & 56.860 & 1412 \\ \hline
\multicolumn{2}{c|}{Horst (same $\nu$)} & 58.100 & 45.773 & \rule{0pt}{2ex} 899 \rule{0pt}{2ex} \\
\multicolumn{2}{c|}{Horst (best $\nu$)} & 57.190 & 56.628 & \rule{0pt}{2ex} 882 \rule{0pt}{2ex} \\
\multicolumn{2}{c|}{Horst+rcca} & 57.236 & 56.856 & \rule{0pt}{2ex} 636 \rule{0pt}{2ex}
\end{tabular}
\caption{Running times, training and test canonical correlations
for a single node Matlab implementation.  ``same $\nu$'' is Horst
run with the same regularization as \RandomizedCCA; this overfits the
test set.  ``best $\nu$'' is the in-hindsight best choice of $\nu$
for generalization.}
\label{tab:ccahypertimes}
\end{minipage}
\end{subtable}
\caption{Europarl results.}
\vspace{-5pt}
\end{figure}

Figure \ref{fig:ccahyperresults} shows the sum of the first 60
canonical correlations found by \RandomizedCCA as the hyperparameters
of the algorithm (oversampling $p$ and number of passes $q$) are
varied.  $\lambda_a$ and $\lambda_b$ are set using the scale-free
parameterization $\lambda_a = \nu \Tr\left( \mat{A}^\top \mat{A}
\right) / d_a$ and $\lambda_b = \nu \Tr \left( \mat{B}^\top \mat{B}
\right) / d_b$, with $\nu = 0.01$.  Figure \ref{fig:ccahyperresults}
indicates that with sufficient oversampling \RandomizedCCA can achieve
an objective value close to that achieved with Horst iteration.  Note in
all cases the solutions found are feasible to machine precision, i.e.,
each projection $\mat{X}_{a,b}$ has (regularized) identity covariance
and the cross covariance $\mat{X}_a^\top \mat{X}_b$ is diagonal.

Table \ref{tab:ccahypertimes} shows single-node running times\footnote{Not
including I/O (all data fits in core) and preprocessing.} and objective
values for \RandomizedCCA with selected values of hyperparameters
and for Horst iteration.  This table indicates that, when iteration
is inexpensive (such as when all data fits in core on a single node),
Horst iteration\footnote{Gauss-Seidel variant with approximate least
squares solves and Gaussian random initializer.} is more efficient
when a high-precision result is desired.  Under these conditions
\RandomizedCCA is complementary to Horst iteration, as it provides
an inexpensive initialization strategy, indicated in the table as
\texttt{Horst+rcca}, where we initialized Horst iteration using the
solution from \RandomizedCCA with $p = 1000$ and $q = 1$.  The overall
running time to achieve the same accuracy, including the time for
computing the initializer, is lower for \texttt{Horst+rcca}.  Furthermore,
the total number of data passes to achieve the same accuracy is reduced
from 120 to 34.

\captionsetup[figure]{font=scriptsize}
\begin{wrapfigure}{r}{0.45\textwidth} 
\vspace{-125pt}
  \begin{center}
    \noindent\makebox[0.45 \textwidth]{%
    \includegraphics[width=0.65\textwidth]{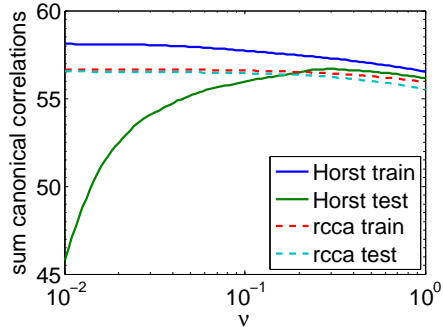}
    }
    \begin{minipage}{0.37 \textwidth}
    \vspace{-185pt}
     \centering
     \captionof{figure}{%
  Effect of $\nu$ on train and test performance.  \RandomizedCCA is run with $q=2$ and $p=2000$.  Horst is run with a budget of 120 data passes.
     }
    \label{fig:nu}
   \end{minipage}
  \end{center}
  \vspace{-110pt}
\end{wrapfigure}
\clearcaptionsetup{figure}
If we view \RandomizedCCA as a learning algorithm, rather than
an optimization algorithm, then the additional precision that Horst
iteration provides may no longer be relevant, as it may not generalize
to novel data.  Alternatively, if sufficiently strong regularization
is required for good generalization the approximations inherent in
\RandomizedCCA are more accurate.  In table \ref{tab:ccahypertimes}
both training and test set objectives are shown.  When Horst in run
with the same regularization as \RandomizedCCA, training objective
is better but test objective is dramatically worse.  By increasing
$\nu$ this can be mitigated, but empirically Horst iteration is more
sensitive to the choice of $\nu$, as indicated in figure \ref{fig:nu}.
This suggests that \RandomizedCCA is providing inherent regularization
by virtue of focusing the optimization on the top range of $\mat{A}^\top
\mat{B}$, analogous to the difference between ridge regression and PCA
regression~\cite{dhillon2013risk}.

\section{Conclusion}

We have presented \RandomizedCCA, a fast approximate CCA solver
which optimizes over the top range of the cross correlation matrix.
\RandomizedCCA is highly amenable to distributed implementation,
delivering comparable accuracy to Horst iteration while requiring far
less data passes.  Furthermore, for configurations where iteration is
not expensive, \RandomizedCCA provides an inexpensive initializer for
Horst iteration.  Finally, when generalization is considered, preliminary
experiments suggest \RandomizedCCA provides beneficial regularization.

\subsubsection*{References}

\begingroup
\renewcommand{\section}[2]{}%
\bibliography{randcca}{}
\endgroup
\bibliographystyle{plain}

\end{document}